%% file: Camera_Ready.tex
\documentclass[10pt,twocolumn,letterpaper]{article}

\usepackage{iccv}              

\usepackage{times}
\usepackage{latexsym}
\usepackage{multirow}
\usepackage{booktabs}
\usepackage{makecell}
\usepackage{subcaption}
\usepackage{threeparttable}
\usepackage{graphicx}
\usepackage{xcolor}
\usepackage{colortbl}
\usepackage{lipsum}
\usepackage{amsmath}
\usepackage{float}
\usepackage{marvosym}


\definecolor{iccvblue}{rgb}{0.21,0.49,0.74}
\usepackage[pagebackref,breaklinks,colorlinks,allcolors=iccvblue]{hyperref}

\def\paperID{1389} 
\def\confName{ICCV}
\def\confYear{2025}

\title{Bootstrapping Grounded Chain-of-Thought in Multimodal LLMs for Data-Efficient Model Adaptation}

\author{%
  Jiaer Xia$^1$ \quad Bingkui Tong$^2$ \quad Yuhang Zang$^3$ \quad Rui Shao$^4$ \quad  Kaiyang Zhou$^1$\textsuperscript{\Letter} \\
  $^1$Hong Kong Baptist University \\
  $^2$Sichuan University \quad $^3$Shanghai AI Lab \quad $^4$Harbin Institute of Technology (Shenzhen) \\
  \url{https://github.com/maifoundations/GCoT}\\
}

\begin{document}
\maketitle
\renewcommand{\thefootnote}{\Letter}
\footnotetext[1]{Corresponding author}

\begin{abstract}
Multimodal Large Language Models (MLLMs) have demonstrated remarkable capabilities in interpreting images using natural language. However, without using large-scale datasets for retraining, these models are difficult to adapt to specialized vision tasks, e.g., chart understanding. This problem is caused by a mismatch between pre-training and downstream datasets: pre-training datasets primarily concentrate on scenes and objects but contain limited information about specialized, non-object images, such as charts and tables. In this paper, we share an interesting finding that training an MLLM with Chain-of-Thought (CoT) reasoning data can facilitate model adaptation in specialized vision tasks, especially under data-limited regimes. However, we identify a critical issue within CoT data distilled from pre-trained MLLMs, i.e., the data often contains multiple factual errors in the reasoning steps. To address the problem, we propose Grounded Chain-of-Thought (GCoT), a simple bootstrapping-based approach that aims to inject grounding information (i.e., bounding boxes) into CoT data, essentially making the reasoning steps more faithful to input images. We evaluate our approach on five specialized vision tasks, which cover a variety of visual formats including charts, tables, receipts, and reports. The results demonstrate that under data-limited regimes our approach significantly improves upon fine-tuning and distillation.
\end{abstract}

\begin{figure}[htbp]
    \centering
    \includegraphics[width=.83\columnwidth]{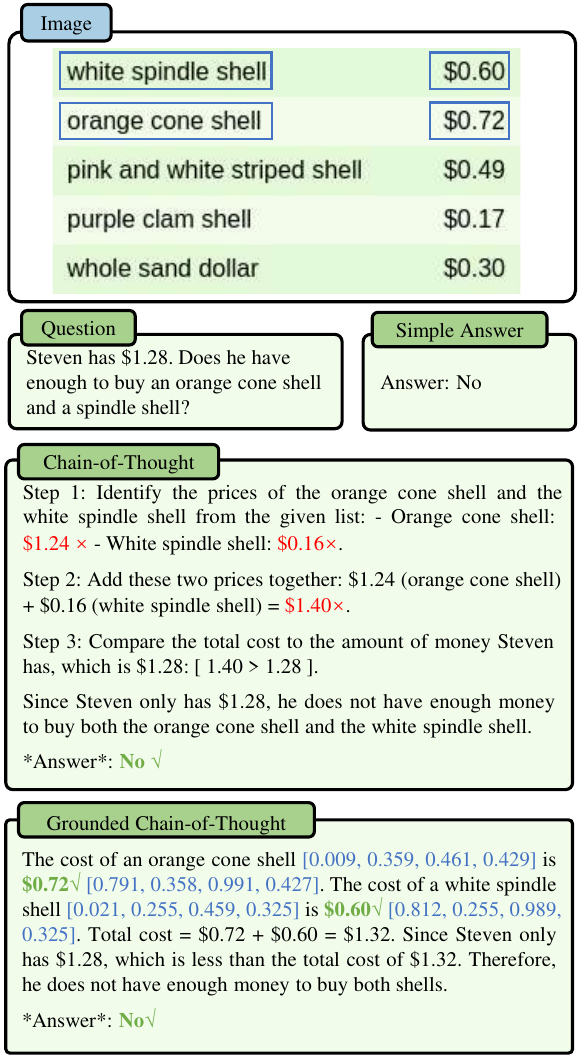}
    \caption{\textbf{Comparison between normal and grounded Chain-of-Thought (CoT).} The normal CoT contains \textcolor{red}{factual errors}, despite giving the \textcolor{PineGreen}{correct} answer. In addition to reasoning, the grounded CoT (GCoT) uses \textcolor{CornflowerBlue}{bounding boxes} to ground key information in the image, which can be self-verified by the model and help minimize factual errors.}
    \label{Fig_1}
\end{figure}

\section{Introduction}

In recent years, Large Language Models (LLMs) have dramatically reshaped the landscape of AI research, including both natural language processing and computer vision. In computer vision, LLMs are typically treated as a decoder; they are combined with a vision encoder in such a way that the output of an LLM is conditioned on the vision encoder's output (namely image features) and a natural language prompt~\cite{chen2024lion, blip2}. Such hybrid models are known as Multimodal Large Language Models (MLLMs) as they can handle both image and text modalities. Although MLLMs excel at interpreting images using natural language, they often fail to adapt to specialized tasks like chart understanding without retraining on large-scale, task-specific datasets~\cite{chartassisstant, chartllama, wu2025grounded}. This is due to a mismatch between pre-training and downstream datasets: MLLMs are mostly pre-trained on object-centric Internet images; as a consequence, MLLMs become inherently weak at handling specialized, non-object images, such as charts and tables.

In this work, we argue that MLLMs that are capable of reasoning can adapt more quickly and efficiently to specialized vision tasks. The intuition is simple: reasoning-capable MLLMs can infer underlying structures, relationships, and logical patterns in data; this allows them to generalize better when it comes to novel data distributions and perform robustly even when exposed to limited training data. As a proof-of-concept, we use Chain-of-Thought (CoT)~\cite{chain} reasoning data distilled from a pre-trained, third-party MLLM to fine-tune our model directly on downstream tasks under few-shot settings. The benchmark covers a wide range of visual formats including charts, tables, receipts, and reports. The average results (shown in Fig.~\ref{main_experiment} (a)) indicate that using around 16 labeled examples the model can beat the zero-shot and fine-tuning baselines by a decent margin. However, by digging into the distilled CoT data, we find a significant limitation: though the final answers may be correct, the intermediate reasoning steps often contain multiple factual errors. Fig.~\ref{Fig_1} shows an example where the intermediate reasoning steps in the CoT data contain wrongly-detected prices: the orange cone shell costs \$0.72 rather than \$1.24, and the white spindle shell costs \$0.60 rather than \$0.16.

To address the limitation in distilled CoT data, we propose Grounded Chain-of-Thought (GCoT), a simple approach that aims to inject grounding information into CoT data with the hope that the reasoning steps become more faithful to input images and therefore the model trained with such data generalizes better. Since it is challenging to collect grounded CoT data, we propose a simple bootstrapping strategy to iteratively bootstrap an MLLM to generate grounding labels and refine them via self-verification. Specifically, we start with an MLLM pre-trained on visual grounding datasets and prompt it to produce bounding boxes for key information (e.g., numbers and objects) extracted from distilled CoT data. Then, we crop images using these bounding boxes and pass the resulting image patches to the model, which performs self-verification by comparing the contents with the pre-extracted key information. After a number of self-verification steps, the grounding information is combined with the CoT data for model fine-tuning. Fig.~\ref{Fig_1} (bottom) shows an example of our grounded CoT where the accurate localization information helps the model identify the correct prices of orange cone shell and white spindle shell.

To evaluate our approach, we build a benchmark that consists of five specialized vision datasets, targeting recognition on charts, tables, receipts, and reports. We compare our approach with several baselines including the zero-shot method, fine-tuning, and distillation (which directly fine-tunes an MLLM with distilled CoT data). The results show that our approach significantly outperforms these baselines under data-limited regimes, demonstrating the effectiveness of grounded CoT data.

In summary, we make the following contributions: 1) We find that CoT data can facilitate model adaptation from generic to specialized vision tasks. We also identify a critical issue in CoT data distilled from pre-trained MLLMs, i.e., the reasoning steps often contain multiple factual errors. 2) To correct factual errors in distilled CoT data, we propose a bootstrapping-based approach that injects self-verified grounding information into CoT to make the reasoning steps more faithful to input images. 3) We provide extensive results on five specialized vision tasks to demonstrate that grounded CoT is the key to enabling data-efficient model adaptation.

\begin{figure*}[t]
    \centering
    \includegraphics[width=\textwidth]{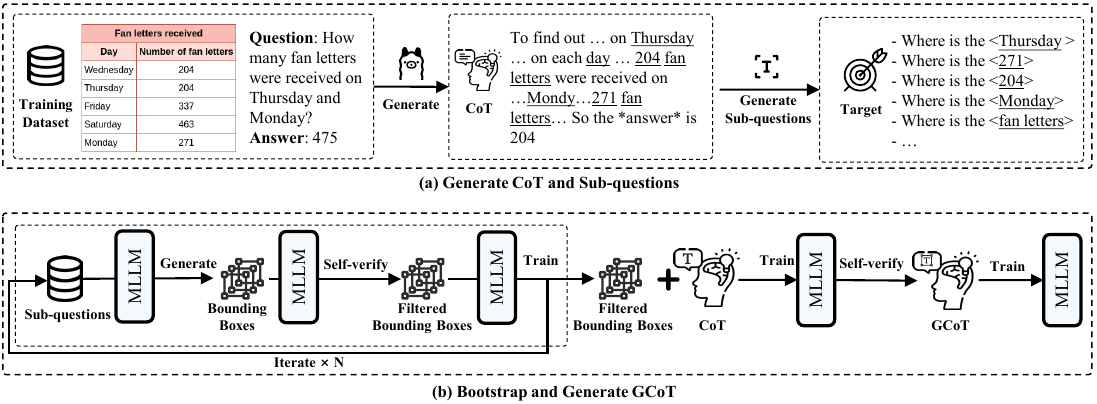}
    \caption{\textbf{Overview of GCoT generation process.} For each sampled training data point, we start by generating a distilled CoT using a third-party model. Key information from the CoT is then extracted to form a set of sub-questions, which are fed into a bootstrapping loop. This loop iteratively generates bounding boxes and filters out the correct ones to improve the model's grounding capability. The finalized filtered bounding boxes are then used to create the GCoT.} \label{Fig_2}
\end{figure*}

\section{Related Work}

\paragraph{Multimodal LLMs}
To endow LLMs with the capability to understand image pixels, the community has conducted numerous studies to integrate LLMs with vision models~\cite{llava1.5, shen2024mome, zhang2025falcon}. The most common approach is to use a well-trained vision model for feature extraction. The image features are then projected onto the text space through fine-tuning a projection layer that connects the vision model with an LLM. Following this practice, LLaVA~\cite{llava} trains the projection layer using data from the CC-3M dataset~\cite{CC3M}, which consists of millions of images and captions crawled from the Internet. Similarly, models like BLIP2~\cite{blip2} and MiniGPT~\cite{minigpt2} further incorporate data from broader sources like LAION400M~\cite{LAION}, Visual Genome~\cite{VG}, and SBU~\cite{SBU} into the training pipeline. It is worth noting that the data widely used for modality alignment is dominated by natural images, which mainly focus on scenes and objects. As a consequence, the model trained with such data becomes weak at recognizing specialized visual formats, such as charts and tables. Our work addresses this problem with a data-efficient model adaptation approach, which allows pre-trained MLLMs to be quickly adapted to specialized vision tasks with few labeled question-answer pairs.

\paragraph{Chart Understanding}
Charts are rich in information and commonly encountered in daily life, playing a crucial role in data visualization across various domains, including business, finance, healthcare, and scientific research. This has fueled a growing interest in developing chart-based LLMs to automate chart interpretation, question answering, and reasoning tasks. Since MLLMs are naturally weak at chart understanding due to lack of exposure to such data during pre-training, many studies have built chart datasets for model adaptation. ChartLLaMA~\cite{chartllama} developed a dataset of 160,000 chart samples. ChartAssistant~\cite{chartassisstant} took a more extensive approach by curating a large-scale dataset containing 39 million chart-text annotated data points. ChartGemma~\cite{chartgemma} developed a chart understanding and reasoning model using 122,857 chart data points. While these datasets have significantly contributed to model adaptation, they come with substantial costs in terms of data collection, annotation, and computational resources. To facilitate model adaptation and reduce the overall production cost, it is imperative to explore data-efficient adaptation methods for fine-tuning MLLMs. Our work provides a timely solution to address this problem.

\paragraph{Chain-of-Thought Reasoning}
The community has observed that CoT reasoning can significantly improve LLMs' performance across a wide range of reasoning tasks~\cite{chain, zero}. By breaking down complex problems into intermediate reasoning steps, CoT enables models to tackle more sophisticated queries that require logical deduction. Meanwhile, existing research has also shown that CoT data is not only beneficial for improving reasoning accuracy but also plays a crucial role in aligning LLMs with desired behaviors and response patterns~\cite{aligning}. One straightforward approach to leveraging CoT in pre-trained LLMs is to prompt them to perform reasoning directly. However, this approach comes with inherent limitations: there is no guarantee that the generated reasoning is always correct, factual, or logically consistent. To address this issue, verification methods have been actively studied in natural language processing, where various techniques have been proposed to check the correctness of CoT reasoning~\cite{verification, selfcheck, training}. However, extending CoT verification to vision-based tasks remains an open question. Our work fills this gap by combining CoT with grounded bounding boxes, which can be verified by a detection-based MLLM.

\section{Methodology}

We propose Grounded Chain-of-Thought (GCoT), an approach that bootstraps an MLLM to generate and refine grounded Chain-of-Thought (CoT). This is motivated by the hypothesis that MLLMs with the ability to perform CoT reasoning can learn new tasks in a faster and more efficient way. A straightforward way is to distill CoT data from a pre-trained MLLM, such as LLaMA 3.2~\cite{llama}. However, the generated CoT data may contain factual inaccuracies in the thinking process, which could negatively impact the training. Our idea is to inject grounding information, i.e., bounding boxes, into CoT, which allows the model to perform self-verification, thereby filtering out incorrect and low-quality CoT data to reduce noisy content.

\input{Table/self_verify}

In general, the GCoT approach consists of three stages: pre-training, bootstrapping, and fine-tuning. First, we pre-train a base MLLM using visual grounding data. This helps the model develop localization ability, i.e., enabling the model to generate bounding boxes that localize key text information in images. Second, given a downstream task with limited training data (e.g., 8 question-answer pairs), we prompt a pre-trained third-party MLLM to generate initial CoT answers, to which we further inject bounding boxes bootstrapped from our model. Finally, we fine-tune our model on the grounded CoT data. See below for more detailed technical designs.

\subsection{Pre-training for Visual Grounding}
We first pre-train a base MLLM, e.g., LLaVA~\cite{llava1.5}, using some visual grounding data. The goal is to enable the model to use bounding box to localize target(s) mentioned in a prompt, such as ``\textit{where is the $<$target$>$}'' where the \textit{$<$target$>$} could be a specific object like chair or a number displayed in the image. In terms of visual grounding data, one could repurpose object detection datasets by simply formulating each detection as a question-answer pair or combining existing VQA datasets that have already provided bounding boxes, such as Flickr30k~\cite{flickr30k} and Visual7W~\cite{visual7w}. In our implementation, we directly adopt the VisCoT-7B model developed by Shao et al.~\cite{viscot}. VisCoT-7B was pre-trained on 10 visual grounding datasets, which cover various domains such as text recognition, general VQA, inforgraphics understanding, and relation reasoning.

\subsection{Generating Grounded Chain-of-Thought}
\paragraph{Initial CoT}
Existing training dataset for chart understanding mostly only contain a question and simple answer, lacking the detailed CoT. With the limited CoT, we first utilize a third-party model (e.g., LLaMA3.2) to generate reasoning process. Given a training dataset $\mathcal{D}=\{Q_i,A_i\}_{i=1}^N$, the third-party model will generate a CoT $C(Q_i)$ for each question $Q_i$.

\paragraph{Bootstrapping}
For each CoT $C(Q_i)$, we use the Natural Language Toolkit (NLTK)~\cite{nltk} to select meaningful nouns and numerical terms as targets. We then construct sub-questions using the template ``\textit{Where is the $<$target$>$?}'', where \textit{$<$target$>$} serves as a placeholder for each target, and such template format is designed to promote the model to generate the bounding box of the target. For each CoT $C(Q_i)$, we generate a set of sub-questions $ \{S_{i,t}\}_{t=1}^{T_i} $, where $T_i$ denotes the total number of targets in $C(Q_i)$, and each sub-question $S_{i,t}$ corresponds to the $t$-th target. Using these sub-questions, we iteratively bootstrap the MLLM to generate grounding information and perform self-verification. In each iteration, the MLLM produces candidate bounding boxes $ \mathcal{B}_{i,t} $ for each sub-question $S_{i,t}$, with the corresponding prompt ``\textit{Please provide the bounding box coordinate of the region.}''. We then crop the area defined by $ \mathcal{B}_{i,t} $, and leverage MLLM itself to process the isolated region, using the prompt ``\textit{The content in this image is:}'' to detect the enclosed content. The identified content is compared to the original target object as a consistency check. Bounding boxes that consistently match the target are retained, while any discrepancies result in the automatic filtering of incorrect proposals. After each iteration, we obtain a set of correct bounding boxes corresponding to the sub-questions:
$$
\mathcal{B}_i = \{ \mathcal{B}_{i,t} \}_{t=1}^{T_i^\prime},
$$
where $\mathcal{B}_{i,t}$ is the correct bounding box for $S_{i,t}$, and $T_i^\prime$ is the number of correct bounding boxes for each CoT $C(Q_i)$. We then use the collected correct bounding boxes to fine-tune the MLLM, enhancing its ability to accurately localize information within images. This iterative process allows the model to identify more bounding boxes in subsequent cycles, progressively refining its skills. Ultimately, the model will develop a robust grounding ability, capable of generating numerous accurate bounding boxes.

\paragraph{Grounded CoT}
After performing the bootstrap process for a pre-determined number of iterations, we combine the CoT with the correct bounding boxes generated in the final iteration to produce the Grounded CoT (GCoT). Specifically, we append the coordinates of the correct bounding boxes directly after the corresponding targets in the CoT. This augmented GCoT is then used to fine-tune the MLLM, enabling it to reason before providing an answer and explicitly output grounding visual information as verifiable evidence. After this fine-tuning, we prompt the MLLM to generate new, high-quality GCoTs in response to the same questions. To guarantee the quality of the data, we prompt the MLLM to generate a batch of GCoTs and then verify the correctness of both the answers and bounding boxes, selecting the correct GCoT. For data augmentation, there will be three verified GCoTs selected for each question.

\subsection{Fine-tuning with Grounded Chain-of-Thought}
As the example show in Tab.~\ref{self-verify}, the self-verification process not only checks GCoT based on results but can also eliminate noisy data from the thinking process through box verification. The bounding box is verified like the process used in the bootstrapping loop, which evaluates the consistency between the contents of the box and the associated target text. We assess its consistency with the target by using MLLM to recognize the box's contents and perform a consistency check. Finally, we retrain the MLLM to further enhance its performance using the newly generated GCoT, which expands the training data while ensuring quality.

\input{Fig/Main_Exp}

\section{Experiments}

\subsection{Datasets}
\label{dataset}
We evaluate our approach, namely Grounded Chain-of-Thought (GCoT), on five publicly available specialized computer vision datasets: ChartQA~\cite{chartqa}, TabMWP~\cite{tabmwp}, SROIE~\cite{sroie}, DVQA~\cite{dvqa}, and TAT-QA~\cite{tat}. These datasets cover a variety of visual formats (i.e., charts, tables, receipts, and reports) across multiple domains, such as finance, science, industry, and mathematics. ChartQA comprises 20,882 charts, including bar, line, and pie charts, paired with 32,719 questions sourced from various sectors like economy and industry. TabMWP features 38,431 grade-level mathematical problems presented in tabular formats, focusing on mathematical reasoning through both free-text and multiple-choice questions. SROIE consists of 1,000 scanned receipt images and focuses on retrieving information from text-heavy document images. DVQA includes bar charts generated from raw data, addressing challenges related to data retrieval and structural comprehension. TAT-QA contains 16,552 questions based on 2,757 hybrid contexts derived from financial reports, with an emphasis on numerical reasoning using image-based tables extracted from PDFs.

\subsection{Evaluation}
We mainly evaluate our approach under data-limited regimes as specialized vision tasks often contain limited labels. Specifically, we train the model using sample sizes of 8, 16, 32, 64, and 128 data points, respectively. To make sure our findings are reliable, we perform three independent random samplings for each sample size and average the results. The standard deviation of each result is also reported. Accuracy (\%) is chosen as the performance metric.

\subsection{Baseline Methods}
Our approach is compared with three baseline methods: 1) \textbf{Zero-shot} directly applies the model in a standard zero-shot setting without any update in the parameters. 2) \textbf{Fine-tuning} fine-tunes the model using the original question-answer pairs provided by the benchmarking datasets. 3) \textbf{Distillation} fine-tunes the model using Chain-of-Thought (CoT) distilled from LLaMA 3.2~\cite{llama}. All these baseline methods use the same backbone as our approach.

\subsection{Implementation Details}
We use VisCoT-7b~\cite{viscot} as the MLLM backbone, which was pre-trained on visual grounding datasets. The model can perform detection based on user prompts, such as \textit{where is the cat}. Our GCoT is based on CoT distilled from LLaMA 3.2~\cite{llama}. All methods that involve fine-tuning are based on LoRA~\cite{lora}. In terms of hyper-parameters, the rank and alpha in LoRA are set to 16 and 32, respectively. AdamW~\cite{adamw} is used as the optimizer, with a learning rate of $2 \times 10^{-4}$. Training is conducted for one epoch. The temperature value is set to 0.8 to encourage diversity when prompting our model to generate GCoT. For each question, we generate 3 GCoT data points.

\subsection{Main Results}

The results are shown in Fig.~\ref{main_experiment}. We summarize our key observations below.

\paragraph{GCoT Consistently Outperforms Baselines}
It is clear that GCoT consistently outperforms all baseline methods across all datasets and training sample sizes. In particular, the green curve, representing GCoT, maintains a clear lead on every benchmarking dataset. This strongly justifies the effectiveness of adding grounding information to CoT. When using only 8 training samples, GCoT beats Zero-shot with a decent margin of around 2\% (in terms of average performance). Compared with training-based methods, i.e., Fine-tuning and Distillation, GCoT consistently outperforms them with significant margins. Even on the most challenging dataset, TAT-QA, which generally exhibits lower accuracy across all methods, GCoT still manages to outperform Fine-tuning and Distillation. The superiority of GCoT becomes more pronounced as the number of training samples increases.

\paragraph{Distillation Improves Over Fine-tuning}
Distillation proves to be a more effective learning approach compared to Fine-tuning. Recall that the main difference between these two methods is that Distillation uses (distilled) CoT data while Fine-tuning relies on the original question-answer pairs. In general, the curves of Distillation stay above the curves of Fine-tuning (except when using low-shot samples on datasets like TabMWP and DVQA). When the sample size increases, the advantage of using CoT data compared to the simple question-answer pairs becomes clearer.

\paragraph{Performance Improves with More Training Samples}
In general, the performance improves across all methods as the number of training samples increases, but the rate of improvement varies significantly. GCoT maintains a healthy upward trend when more training samples are used. Distillation also has an upward trend in most cases except on ChartQA where the increasing momentum fades away when transitioning from 64 to 128 training samples. This is likely to be caused by the errors in distilled CoT data. When it comes to Fine-tuning, the results are drastically different: the accuracy remains relatively unchanged when more training data is available on ChartQA and TabMWP; this suggests that simple question-answer pairs have limitations and as a result the model does not gain new knowledge with more data. It is worth noting that on TabMWP the accuracy of Fine-tuning even decreases in the 128-sample-size case, which suggests that model overfits the small training data.

\subsection{Ablation Studies and Analyses}

\input{Table/self-verification_analysis}

\paragraph{Self-verification of GCoT}
To evaluate the impact of self-verification in GCoT, we conducted an ablation study with three configurations: (1) using augmentation and box verification, (2) without augmentation (single GCoT per sample), and (3) without box verification (select the augmented data by verifying only the final answer). The results, shown in Tab.~\ref{self-verification_analysis}, reveal the significant impact of both augmentation and box verification mechanisms on the model’s performance. When augmentation is disabled, performance drops by an average of 1.63 to 2.36 percentage points across various sample sizes. Without box verification, the accuracy drops even further, with performance reductions ranging from 5.94 to 10.36 percentage points, which indicates that relying solely on final answer checks is insufficient for filtering out quality CoT data and noisy CoT data can significantly impair alignment performance. These findings highlight the importance of generating multiple CoT annotations and verifying their correctness through box verification to ensure robust performance.

\begin{figure}[htbp]
  \includegraphics[width=0.45\textwidth]{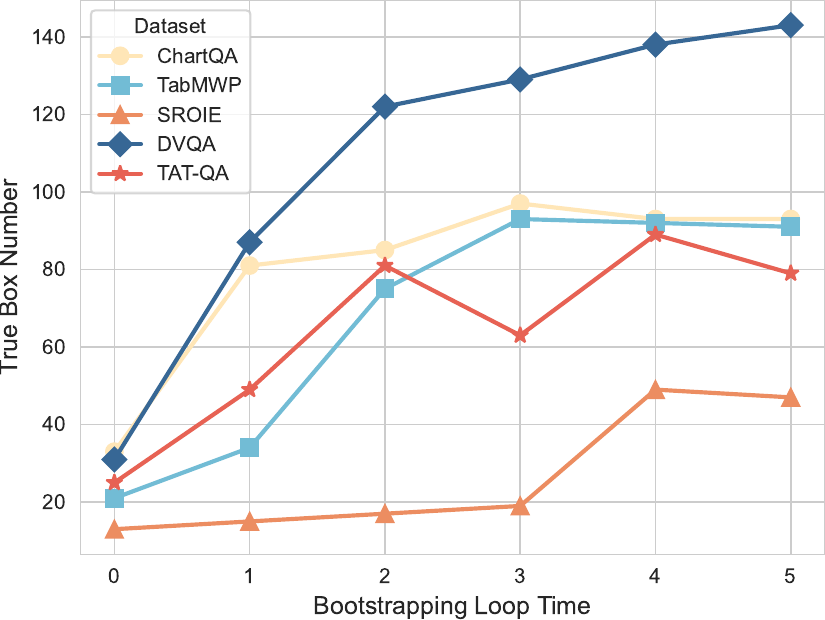}
  \caption{Result of analyzing the impact of bootstrapping loop time.}
  \label{loop}
\end{figure}

\paragraph{Bootstrapping Loop Time}
The results presented in Fig.~\ref{loop} demonstrate the impact of increasing the bootstrapping loop time on the number of true bounding boxes generated. As the bootstrapping loop time increases, the number of true boxes steadily rises for all datasets. This indicates that the iterative process of generating and refining bounding boxes through multiple loops leads to more accurate and reliable outputs. Notably, datasets such as ChartQA and TabMWP show substantial improvements in true box generation, while TAT-QA and DVQA exhibit more gradual gains. This demonstrates the effectiveness of bootstrapping in enhancing the model's ability to generate high-quality bounding boxes over successive iterations, providing an efficient method to obtain grounded information without the need for labeling.

\input{Table/distillation_analysis}

\input{Table/Data_Visualization}
\paragraph{Distillation Choice}
Finally, we analyze the impact of using different models for distilling CoT. Tab.~ \ref{distillation_analysis} shows the performance of GCoT when distilled using various language models, including LLaMA 3.2, Claude 3.5, GPT-4o, Qwen2-VL, and Gemini 1.5-Pro. The results indicate that GCoT consistently outperforms the normal distillation approach, achieving average improvements ranging from +1.54 to +5.62 percentage points across different models. The best results were observed when distilling from Claude 3.5 and Qwen2-VL, where GCoT improves by +5.62 and +4.95 percentage points, respectively. These results indicate that the GCoT is not restricted to any specific source model for distillation, allowing it to receive input from a variety of models without impacting its performance. 

An example of different source distillation data is presented in Tab.~\ref{Distillation}. Overall, the distillation data provided by each model exhibits significantly different formats. This leads to very unstable training when directly applied to such an uncertain data format, as evidenced by the high variance of 2.37 in the average results from Tab.~\ref{distillation_analysis}. In contrast, GCoT alleviates format differences through self-generated data, resulting in more stable training with a variance of only 0.91. At the same time, when delving into the details of each distillation data, each piece of data may have issues related to the thought process, answers, or formatting to varying degrees, which can be harmful to the model's training. Thus, by addressing formatting and filtering errors, our GCoT can derive performance improvements from the distillation data of various sources. Examples of corresponding GCoT data for different sources can be found in the supplementary material.

\section{Conclusion}
To the best of our knowledge, this study is the first to study MLLMs in data-limited settings for specialized vision tasks. Our research, conducted across five distinct datasets, reveals that training models solely with question-answer pairs fails to significantly enhance their generalization capabilities, particularly when only a limited number of labeled examples are available. Although integrating CoT reasoning into the training process results in some performance improvements, the presence of factual inaccuracies within the CoT constrains the model's ability to learn broadly applicable knowledge. To overcome this limitation, we introduce GCoT, which essentially injects self-verified grounding information into the CoT. This approach enables the model to learn from higher-quality reasoning data, thereby capturing more generalizable patterns. As MLLMs continue to grow in size, efficient adaptation becomes essential for their democratization and large-scale deployment. We hope this work provides valuable insights and inspires future research to address these practical challenges in enhancing model generalization under constrained data conditions.

\section{Limitations and Future Work}
In terms of limitations, our approach requires examining the CoT reasoning process based on the content within bounding boxes. For abstract objects like lines or icons, providing accurate bounding boxes and verifying their content is challenging. This restricts the method to image formats that are mainly rich in text and numbers, such as those used in chart analysis. Meanwhile, our model relies on existing external models to provide corresponding CoT data for bootstrapping, which greatly limits the training of the method and makes it difficult to scale up. In future work, we can try to leverage reinforcement learning approaches~\cite{xia2025visionary} to enhance the model's CoT ability and further inject grounding ability.

\section{Acknowledgments}
This project was in part supported by Hong Kong Research Grants Council Early Career Scheme (No. 22200824), and National Natural Science Foundation of China (Grant No. 62306090).

{
    \small
    \bibliographystyle{ieeenat_fullname}
    \bibliography{main}
}

\clearpage
\appendix
\section{Appendix}
\label{sec:appendix}

\subsection{Evaluation of Generalization Ability}
Theoretically, our approach can stem the generalizability from clearer and better alignment training data. Compared to the original data, which only includes simple answer annotations, CoT data provides the detailed underlying structures, relationships, and logical patterns within the data. This enables clearer alignment and understanding, rather than merely remembering the simple answer, thereby capturing more generalizable patterns. Additionally, the proposed GCoT provides higher-quality reasoning data by eliminating the noisy information that may exist in the distilled CoT data. To better demonstrate generalizability, we conducted a cross-model evaluation, as shown in Tab.~\ref{tab:generalization}. The results indicate that the model trained on ChartQA is capable of generalizing improvements to other datasets. Notably, our proposed GCoT method shows even greater enhancements.

\begin{table}[h]
\centering
\caption{The model is trained on 128 ChartQA datasets and demonstrates strong generalization capabilities across different datasets with proposed GCoT method.}
\label{tab:generalization}
\begin{tabular}{@{}lccc@{}}
\toprule
Method   & ChartQA & TabMWP    & DVQA  \\ \midrule
Zero-shot& 10.12   & 19.60     & 13.95 \\
CoT      & 20.44   & 21.01     & 12.34 \\
GCoT     & 25.18   & 25.55     & 15.73 \\ \bottomrule
\end{tabular}
\end{table}

\subsection{Evaluation of Chain-of-Thought Accuracy}
To address the scarcity of detailed annotations for intermediate reasoning steps and to more intuitively validate our process verification approach, we manually examined 64 ChartQA CoT samples, as shown in Table~\ref{tab:accuracy}. The results support the findings from our ablation study: although the distilled CoT contains certain errors, incorporating box verification substantially reduces these mistakes, resulting in more accurate training data. Moreover, if process verification is omitted and only the final answers are checked, the self-augmented data exhibit a higher error rate. In contrast, GCoT effectively alleviates this issue, yielding significantly more reliable training data.

\begin{table}[h]
\centering
\caption{Box-based Verification can effectively eliminate errors and provide more accurate training samples.}
\label{tab:accuracy}
\begin{tabular}{lcccc}
\hline
         & CoT   & GCoT    & w/o Box    & GCoT+Aug      \\ \hline
Acc/All  &49/64  & 57/64   & 33/64      & 162/192       \\ \hline
\end{tabular}
\end{table}

\subsection{Visualization}
This section presents illustrative examples to clarify the details of GCoT. We demonstrate the self-verified GCoT with various sources of distillation data and different datasets, highlighting the improvements brought about by GCoT.
\subsubsection{Different source of Distillation CoT}
\label{sec:cot}
\begin{figure*}[t]
  \includegraphics[width=0.8\textwidth]{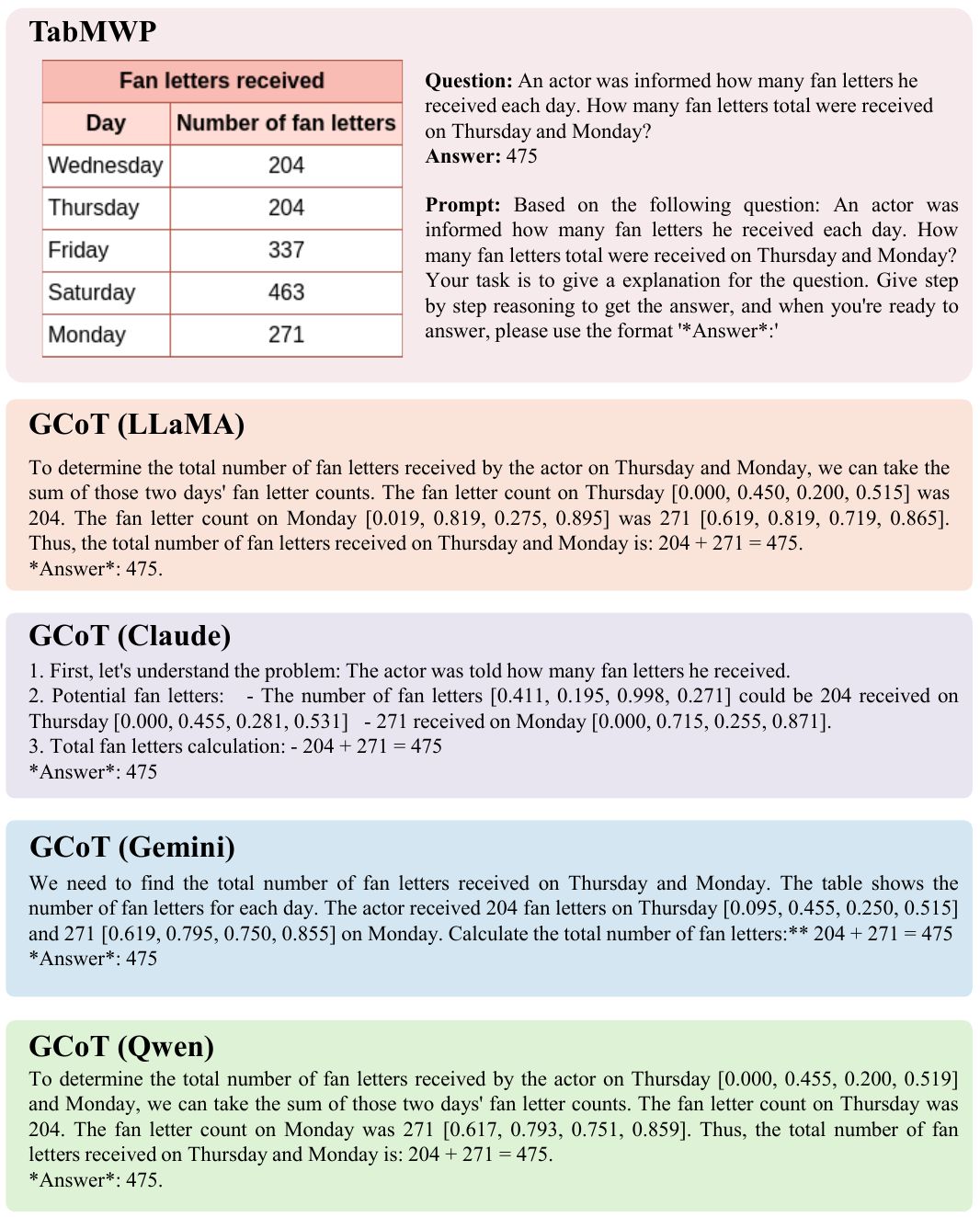}
  \centering
  \caption{Self-generated GCoT data eliminating external language discrepancies and aligning closely with the model's linguistic tendencies.}
  \label{GCoT}
\end{figure*}

In the main paper, we have shown some examples to compare distillation CoT data from different sources and analyze why their performance varies significantly under training. Additionally, we will contrast these with our self-generated GCoT data, with the results illustrated in Fig.~\ref{GCoT}. In contrast to the distillation CoT, the GCoT markedly reduces inaccuracies in the reasoning process, resulting in higher-quality training data.

\subsubsection{Chart Understanding Task Case}
\begin{figure*}[t]
  \includegraphics[width=0.8\textwidth]{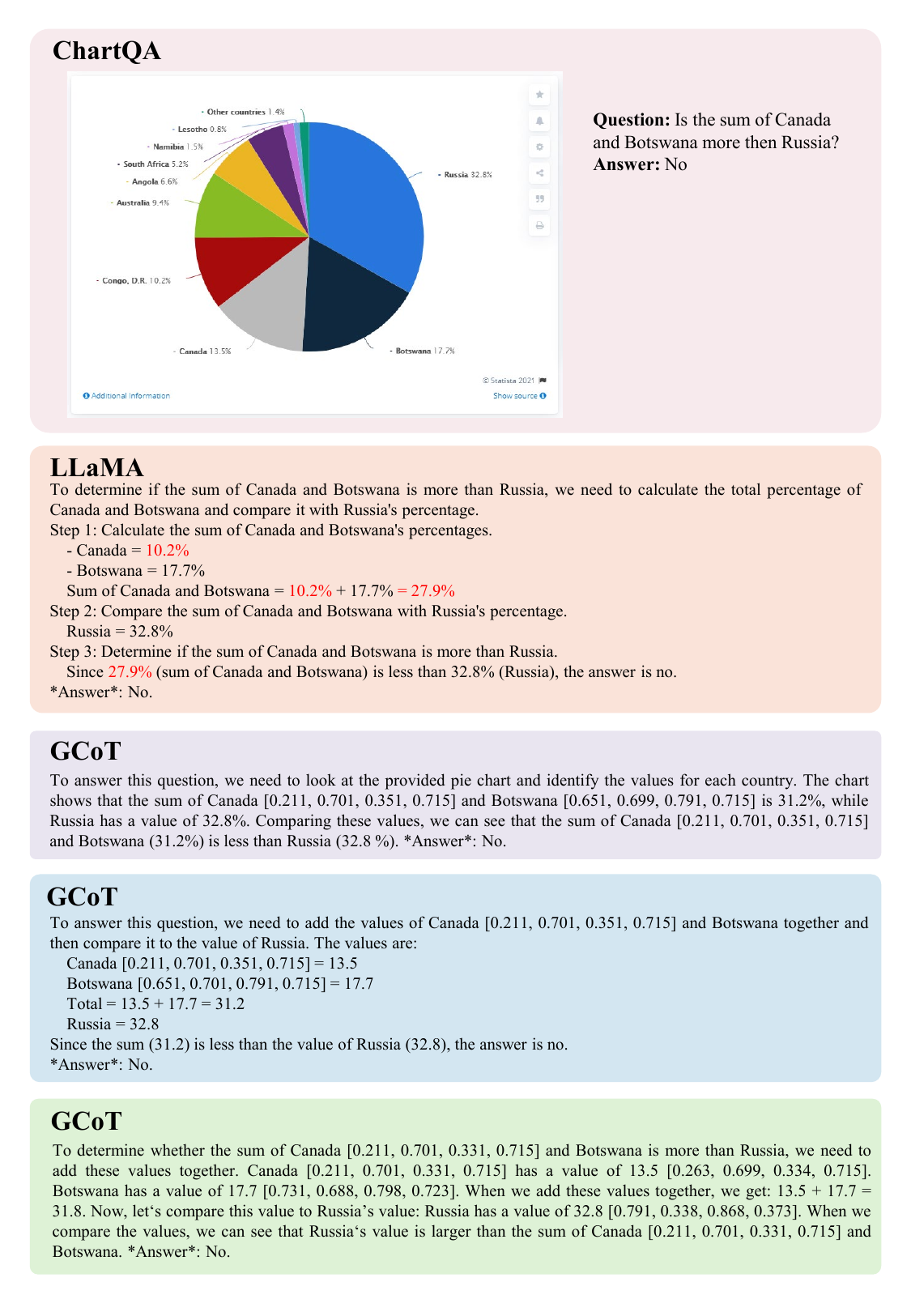}
  \centering
  \caption{Example of GCoT applied to ChartQA dataset, illustrating its ability to generate reliable and diverse training CoT data.}
  \label{ChartQA}
\end{figure*}

\begin{figure*}[t]
  \includegraphics[width=0.8\textwidth]{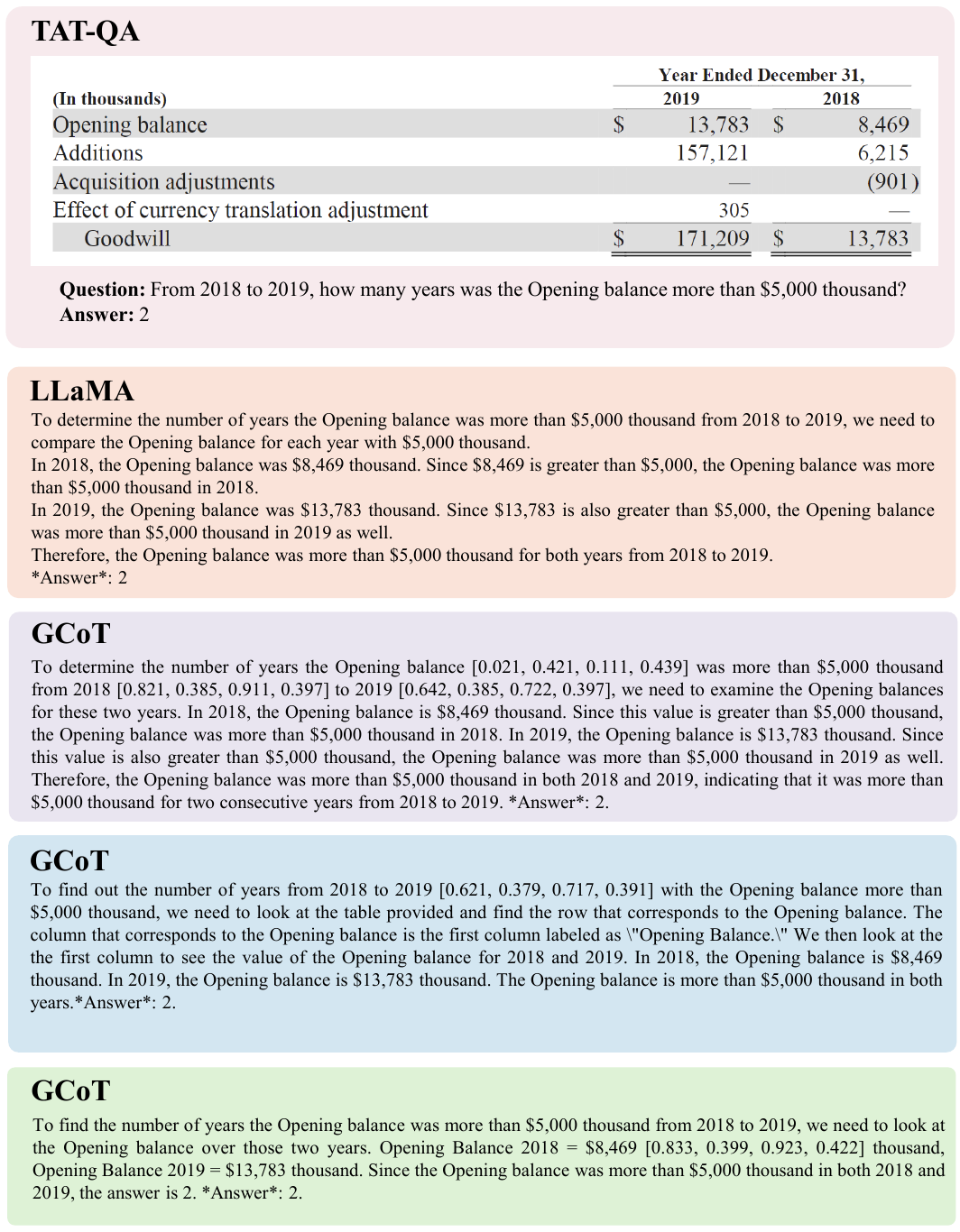}
  \centering
  \caption{Example of GCoT applied to TAT-QA dataset, illustrating its ability to generate reliable and diverse training CoT data.}
  \label{TAT}
\end{figure*}

\begin{figure*}[t]
  \includegraphics[width=0.8\textwidth]{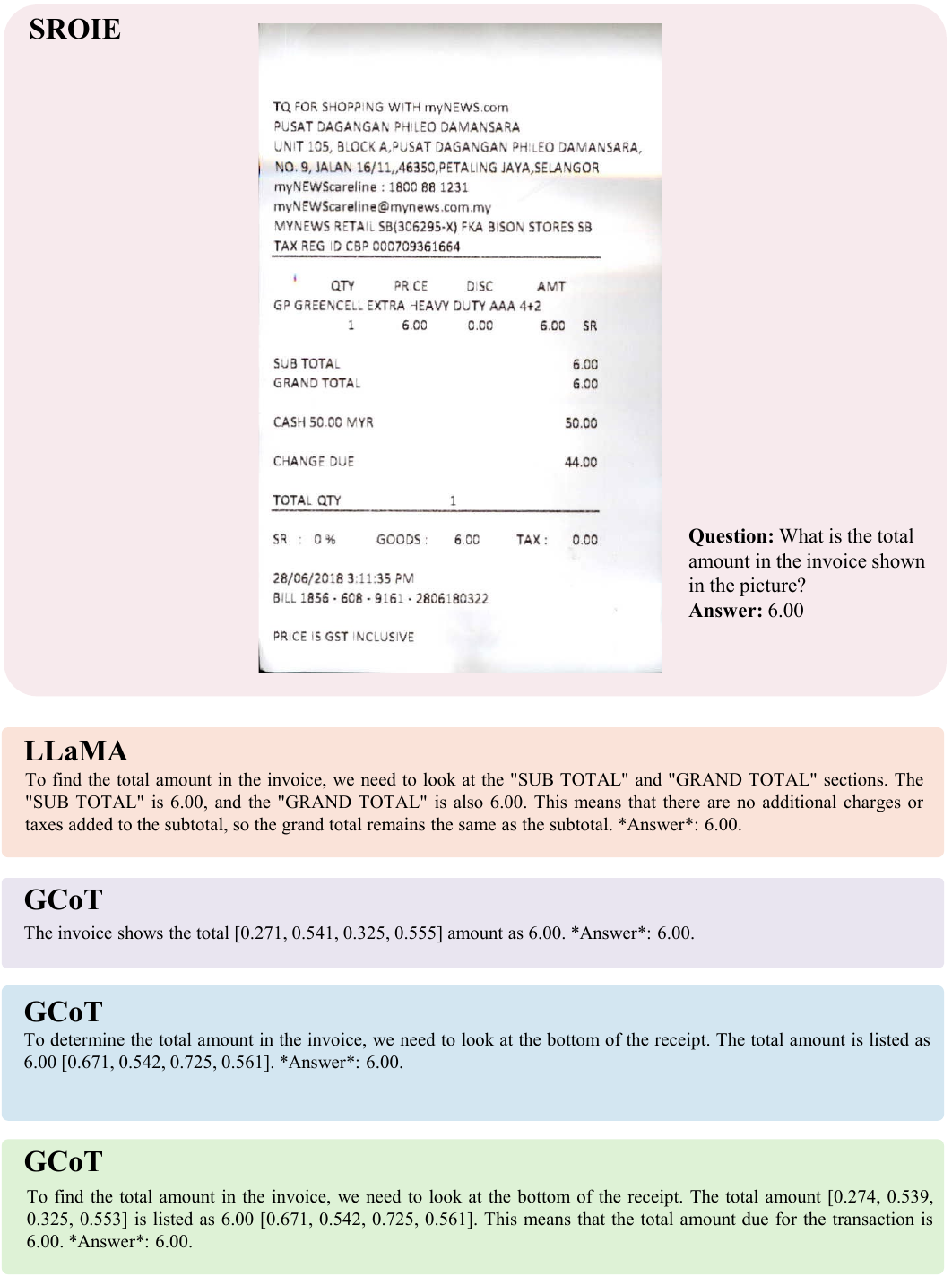}
  \centering
  \caption{Example of GCoT applied to SROIE dataset, illustrating its ability to generate reliable and diverse training CoT data.}
  \label{SROIE}
\end{figure*}

Figs.~\ref{ChartQA}–\ref{SROIE} illustrate GCoT's adaptability across different datasets in the context of chart understanding tasks. Firstly, in comparison to the distillation data, GCoT significantly reduces inaccuracies and standardizes the format. Furthermore, GCoT is capable of generating a diverse range of CoT data while maintaining high accuracy. These examples highlight GCoT's ability to produce reliable and varied datasets through self-generation. This capability not only expands the originally limited datasets but also enhances training efficiency, leading to improved performance across diverse tasks.

\end{document}

%% file: Table/self_verify.tex
\begin{table*}[t]
\centering
\begin{tabular}{@{}m{5.2cm}@{}m{12cm}@{}}
\toprule
\multicolumn{2}{l}{\textbf{Self-verification Example}} 
\\ \midrule
\multicolumn{2}{l}{\small Question: Felix wants to buy 2 kilograms of beef sauce and 2 kilograms of marinara sauce. How much will he spend? Answer: 6.04}
\\ \midrule
\includegraphics[width=5cm]{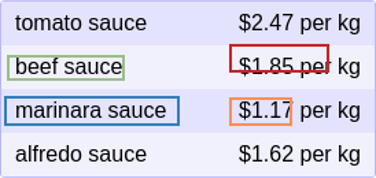} & \footnotesize \textbf{Verified Positive Sample}

To find the total cost of 2 kilograms of beef sauce \textcolor{YellowGreen}{[0.021, 0.411, 0.331, 0.475]} and 2 kilograms of marinara sauce \textcolor{RoyalBlue}{[0.011, 0.521, 0.475, 0.595]}, we need to multiply the prices of each type of sauce by its quantity. The price of beef sauce is \$1.85 \textcolor{Maroon}{[0.611, 0.381, 0.875, 0.455]} per kilogram, so we multiply it by 2 to get \$3.70. The price of marinara sauce is \$1.17 \textcolor{orange}{[0.611, 0.522, 0.775, 0.595]} per kilogram, so we multiply it by 2 to get \$2.34. Total cost = \$3.70 + \$2.34 = \$6.04. *Answer*: \textcolor{PineGreen}{6.04}.                                                                                   \\ \midrule
\includegraphics[width=5cm]{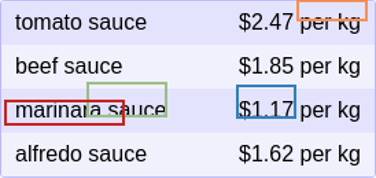} & \footnotesize 
\textbf{Verified Negative Sample}

To find out how much Felix will spend on 2 kilograms of beef sauce \textcolor{YellowGreen}{[0.231, 0.485, 0.445, 0.575]}, we need to multiply the price per kilogram by the number of kilograms he is buying. The price per kilogram of beef sauce is \$1.17 \textcolor{RoyalBlue}{[0.631, 0.492, 0.787, 0.577]}. The total cost of 2 kilograms of beef sauce would be: \$1.17 * 2 = \$2.34. To find out how much Felix will spend on 2 kilograms of marinara \textcolor{Maroon}{[0.011, 0.531, 0.331, 0.595]} sauce, we need to multiply the price per kilogram by the number of kilograms he is buying. The price per kilogram of marinara sauce is \$1.62 \textcolor{orange}{[0.791, 0.265, 0.975, 0.319]}. The total cost of 2 kilograms of marinara sauce would be: \$1.62 * 2 = \$3.24. Total cost = \$2.34 + \$3.24 = \$5.58. *Answer*: \textcolor{Red}{5.58}.  \\
\bottomrule
\end{tabular}
\caption{Example from the self-verify process comparing positive and negative samples. Both the final answer and the reasoning process have been verified to ensure the accuracy of the selected sample.}
\label{self-verify}
\end{table*}

%% file: Fig/Main_Exp.tex
\begin{figure*}[t] 
    \centering 
    \begin{subfigure}[t]{0.315\textwidth}
        \includegraphics[width=\textwidth]{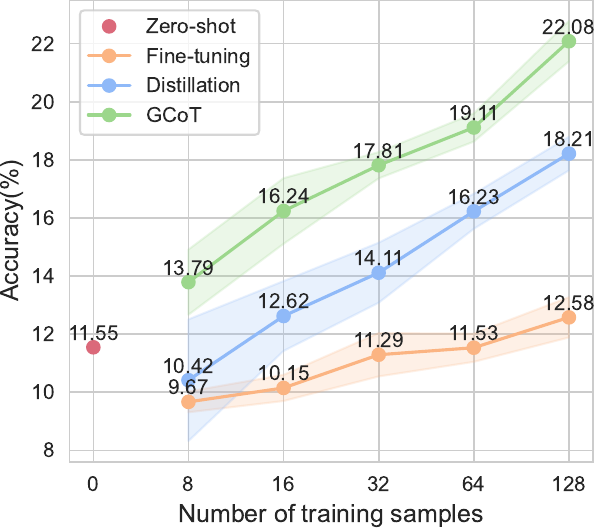}
        \caption{Average}
        \label{main_experiment_average}
    \end{subfigure}
    \begin{subfigure}[t]{0.325\textwidth}
        \includegraphics[width=\textwidth]{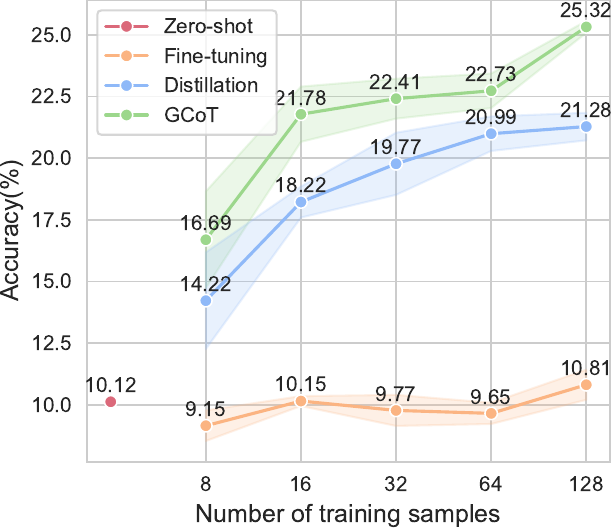}
        \caption{ChartQA}
        \label{main_experiment_chartqa}
    \end{subfigure}
    \begin{subfigure}[t]{0.315\textwidth}
        \includegraphics[width=\textwidth]{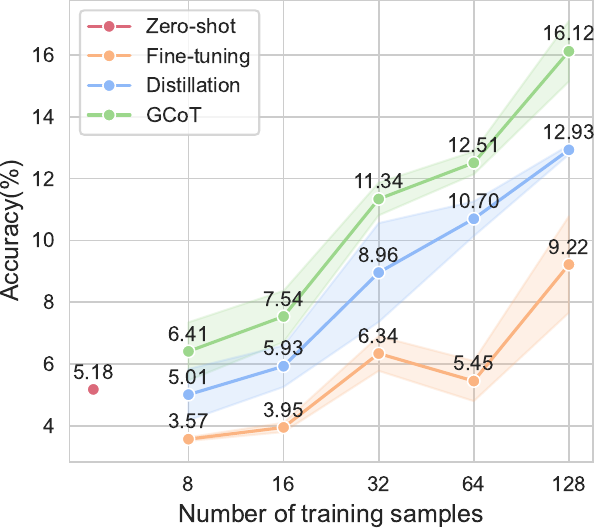}
        \caption{SROIE}
        \label{main_experiment_sroie}
    \end{subfigure}
    \begin{subfigure}[t]{0.32\textwidth}
        \includegraphics[width=\textwidth]{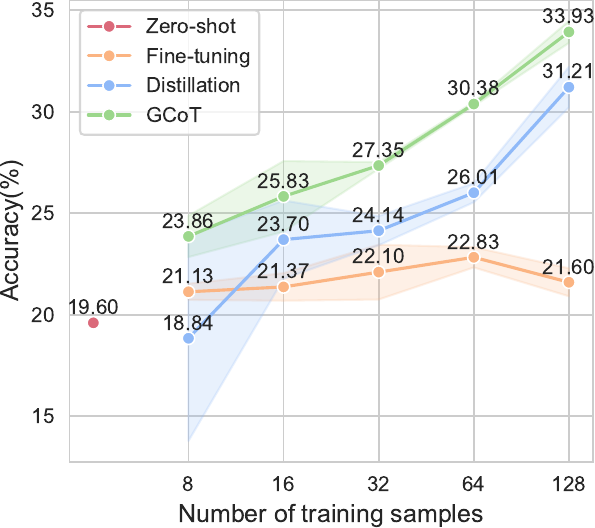}
        \caption{TabMWP}
        \label{main_experiment_tabmwp}
    \end{subfigure}
    \begin{subfigure}[t]{0.32\textwidth}
        \includegraphics[width=\textwidth]{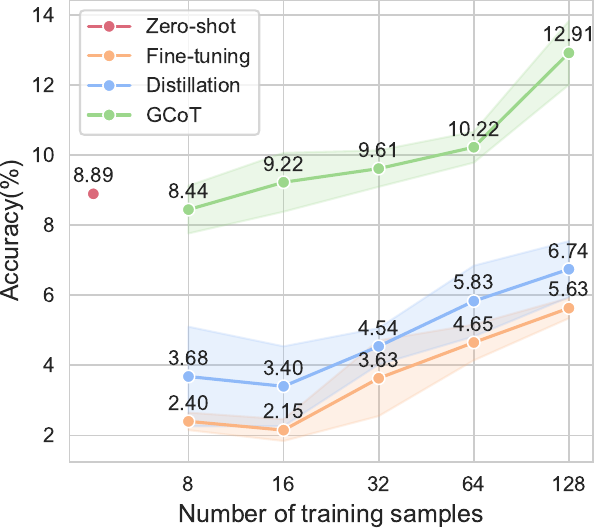}
        \caption{TAT-QA}
        \label{main_experiment_tatqa}
    \end{subfigure}
    \begin{subfigure}[t]{0.32\textwidth}
        \includegraphics[width=\textwidth]{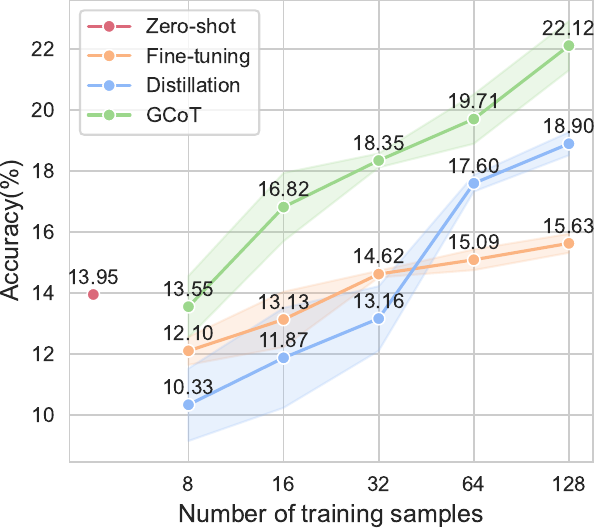}
        \caption{DVQA}
        \label{main_experiment_dvqa}
    \end{subfigure}
    
    \caption{\textbf{Main results on the five specialized vision datasets.}
    Fine-tuning means directly fine-tuning the model with simple question-answer pairs. Distillation means training the model with LLaMA distilled CoT data. All training-based methods are based on the same backbone and the LoRA method.
    Overall, the performance of GCoT surpasses the fine-tuning and distillation methods across all datasets and sample sizes. The shadow areas represent variance.}
    \label{main_experiment}
\end{figure*}

%% file: Table/self-verification_analysis.tex
\begin{table}[htbp]
\centering
\resizebox{0.47\textwidth}{!}{%
\begin{tabular}{@{}cccc@{}}
\toprule
Samples & GCoT              & w/o Augmentation & w/o Box Verification \\ \midrule
8      & \textbf{23.86$_{\pm1.02}$} & 21.56$_{\pm1.91}$ \color[HTML]{CB0000}(-2.30)       & 17.92$_{\pm4.88}$ \color[HTML]{CB0000}(-5.94)    \\
16     & \textbf{25.83$_{\pm1.74}$} & 23.97$_{\pm1.88}$ \color[HTML]{CB0000}(-1.86)       & 19.43$_{\pm2.35}$ \color[HTML]{CB0000}(-6.40)    \\
32     & \textbf{27.35$_{\pm0.17}$} & 25.62$_{\pm0.62}$ \color[HTML]{CB0000}(-1.73)       & 20.96$_{\pm1.56}$ \color[HTML]{CB0000}(-6.39)    \\
64     & \textbf{30.38$_{\pm0.10}$} & 28.75$_{\pm0.29}$ \color[HTML]{CB0000}(-1.63)       & 21.11$_{\pm1.38}$ \color[HTML]{CB0000}(-9.27)    \\
128    & \textbf{33.93$_{\pm0.54}$} & 31.57$_{\pm0.47}$ \color[HTML]{CB0000}(-2.36)       & 23.57$_{\pm2.21}$ \color[HTML]{CB0000}(-10.36)    \\ \bottomrule
\end{tabular}%
}
\caption{Ablation studies examining the components of GCoT with TabMWP dataset. w/o Augmentation means only generating single GCoT data for each sample and w/o Box Verification means only the final answer of generated CoT is verified.}
\label{self-verification_analysis}
\end{table}

%% file: Table/distillation_analysis.tex
\begin{table}[t]
\centering
\resizebox{0.45\textwidth}{!}{%
\begin{tabular}{@{}lcc@{}}
\toprule
Source        & Distillation    & GCoT     \\ \midrule
LLaMA3.2      & 24.14$_{\pm0.72}$ & \textbf{27.35$_{\pm0.17}$ \color[HTML]{009901}(+3.21)} \\
Claude3.5     & 19.84$_{\pm0.49}$ & \textbf{25.46$_{\pm0.49}$ \color[HTML]{009901}(+5.62)} \\
GPT-4o        & 25.91$_{\pm1.80}$ & \textbf{27.46$_{\pm0.22}$ \color[HTML]{009901}(+1.54)} \\
Qwen2-VL      & 21.31$_{\pm0.60}$ & \textbf{26.25$_{\pm0.54}$ \color[HTML]{009901}(+4.95)} \\
Gemini1.5-Pro & 22.70$_{\pm1.14}$ & \textbf{26.20$_{\pm0.95}$ \color[HTML]{009901}(+3.50)} \\ \midrule
\rowcolor[HTML]{EFEFEF} 
Average       & 22.78$_{\pm2.37}$ & \textbf{26.54$_{\pm0.91}$ \color[HTML]{009901}(+3.76)} \\ \bottomrule
\end{tabular}%
}
\caption{Ablation on distillation model choice.}
\label{distillation_analysis}
\end{table}

%% file: Table/Data_Visualization.tex
\begin{table*}
\centering
\begin{tabular}{@{}m{4cm}@{}m{12cm}@{}}
\toprule
\multicolumn{2}{l}{\textbf{Distillation Data Example}} 
\\ \midrule
\centering
\includegraphics[width=3.5cm]{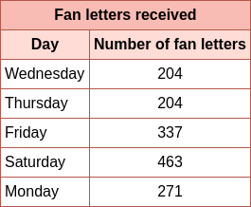} & \footnotesize \textbf{Question:} An actor was informed how many fan letters he
received each day. How many fan letters total were received
on Thursday and Monday?

\textbf{Prompt:} Based on the following question: An actor was
informed how many fan letters he received each day. How
many fan letters total were received on Thursday and Monday?
Your task is to give a explanation for the question. Give step
by step reasoning to get the answer, and when you're ready to
answer, please use the format '*Answer*:'

\textbf{Answer:} 475    
\\ \midrule
\multicolumn{2}{l}{\begin{tabular}[c]{@{}m{16cm}@{}m{}@{}}\footnotesize
\textbf{LLaMA Distillation Data}

To find the total number of fan letters received on Thursday and Monday, we need to add the number of fan letters received on each day.

Step 1: Identify the number of fan letters received on Thursday. According to the table, the actor received 204 fan letters on Thursday.

Step 2: Identify the number of fan letters received on Monday. According to the table, the actor received \textcolor{red}{274 fan letters on Monday.}

Step 3: Add the number of fan letters received on Thursday and Monday. \textcolor{red}{Total = 204 + 274 = 478}

*Answer*: 475
\end{tabular}}
\\ \midrule
\multicolumn{2}{l}{\begin{tabular}[c]{@{}m{16cm}@{}m{}@{}}\footnotesize
\textbf{Gemini Distillation Data}

1. **Identify the relevant data:** The table shows how many fan letters an actor received each day and on Thursday and Monday.

2. **Extract the data:** The table shows the number of \textcolor{red}{fan letters received each day: 204.}

The table also shows the number of fan letters received on \textcolor{red}{Thursday and Monday: 271.}

3. **Add the values:** The total number of fan letters received on Thursday and Monday is 271 + 204 = 475.

*Answer*: 475
\end{tabular}}
\\ \midrule
\multicolumn{2}{l}{\begin{tabular}[c]{@{}m{16cm}@{}m{}@{}}\footnotesize
\textbf{Claude Distillation Data}

Okay, let's break this down step-by-step:

1. The question states that an actor was informed how many fan letters he received each day.

2. The provided table shows the number of fan letters received on each day of the week.

3. The question specifically asks for the total number of fan letters received on Thursday and Monday.

4. From the table, we can see that on Thursday, \textcolor{red}{the actor received 214 fan letters.}

5. On Monday, the actor received 271 fan letters.

6. To get the total number of fan letters received on Thursday and Monday, we add these two numbers together.

*Answer*: 475 fan letters were received on Thursday and Monday.
\end{tabular}}
\\ \midrule
\multicolumn{2}{l}{\begin{tabular}[c]{@{}m{16cm}@{}m{}@{}}\footnotesize
\textbf{Qwen Distillation Data}

To find the total number of fan letters received on Thursday and Monday, we need to add the number received on each of those days.

From the table:

- Thursday: 204 fan letters

- Monday: 271 fan letters

Now, we add these two numbers together: 204 (Thursday) + 271 (Monday) = 475 fan letters

\textcolor{red}{So, the total number of fan letters received on Thursday and Monday is 475.}
\end{tabular}}
\\
\bottomrule
\end{tabular}
\caption{Example of Distillation CoT data from different sources, showing variations in language patterns and errors in reasoning.}
\label{Distillation}
\end{table*}